%% file: iclr2026_conference.tex
\documentclass{article} % For LaTeX2e
\usepackage{iclr2026_conference,times}

\input{math_commands.tex}

\usepackage{hyperref}
\usepackage{url}

 \usepackage{graphicx}
 \usepackage{booktabs} 

\title{Evolutionary Two-Stage Hyperparameter \\ Optimization Strategies for \\ Physics-Informed Neural Networks}

\author{Fedor Buzaev$^1$, Dmitry Efremenko$^1$, Egor Bugaev$^1$, Andrei Ermakov$^{1,2}$, \\
\textbf{Denis Derkach$^1$, Daria Pugacheva$^{* 1,2}$, Fedor Ratnikov}\thanks{Equal advising.}~$\,^1$ \\
$^1$HSE University, $^2$AXXX
}

\iclrfinalcopy % Uncomment for camera-ready version, but NOT for submission.
\begin{document}

\maketitle

\begin{abstract}
Physics-Informed Neural Networks (PINNs) solve Partial Differential Equations (PDEs) by embedding physical laws into neural network training. However, their performance suffers from unstable convergence, training plateaus, and strong sensitivity to architectural and optimization hyperparameters due to the highly non-convex and multi-term structure of the physics-informed loss. In this setting, the outer-loop hyperparameter search is a noisy and black-box optimization problem over heterogeneous parameters, where classical local or gradient-based strategies are easily trapped in suboptimal regions. Evolutionary algorithms, with their population-based exploration and ability to handle mixed, non-differentiable search spaces, provide a more robust mechanism for discovering promising configurations. We propose and investigate a two-stage approach based on evolutionary algorithms that combines exploration and exploitation parts of PINNs training to improve solution accuracy and robustness under fixed computational budgets. In the first stage, we perform low-fidelity training runs with truncated epochs to rapidly screen candidate configurations, treating hyperparameter selection as a black-box outer-loop problem. In the second stage, only the most promising candidates are fully trained with standard gradient-based optimizers to refine the solution. Evaluated on three popular problems, namely Advection, Klein–Gordon and Helmholtz equations, our method consistently outperforms standard training and achieves significantly lower mean error within constrained computational resources.
\end{abstract}

\section{Introduction}
Physics-Informed Neural Networks (PINNs) solve PDEs by minimizing a loss that includes residuals of the governing equations~\citep{raissi2019physics}. They have been applied in fluid dynamics, heat transfer, and quantum mechanics. Despite this promise, PINNs training often stalls, diverges, or converges to suboptimal minima, with robustness poorly understood~\citep{NEURIPS2021_df438e52,Wang2022,Chuprov2024, Buzaev2023, Buzaev2024}. To compete with classical solvers like finite differences~\cite{Mao2020,grossmann2023}, PINNs must be made more efficient and stable.

A central obstacle is sensitivity to hyperparameters. Architectural choices and optimization settings can drastically alter convergence behavior, causing training plateaus or convergence to poor local minima. This sensitivity is particularly challenging because hyperparameter tuning is naturally bilevel: for each hyperparameter configuration, an inner gradient-based optimizer trains the network weights, while an outer procedure must select the hyperparameters that yield the best final solution quality. In practice, this outer-loop objective is a computationally expensive black box, since model quality is only observed after executing a training run. Moreover, the loss landscape of PINNs has been compared to that of standard deep neural networks (DNNs), revealing a significantly higher density of local minima \cite{arxiv:2501.06572,Wang2022}. This increased multimodality is cited as a primary reason why PINN training typically requires more iterations to converge and is prone to suboptimal solutions. Consequently, classical local strategies may prematurely commit to suboptimal regions of the search space.

These considerations motivate the use of population-based evolutionary optimization in the outer loop. Evolutionary algorithms can search effectively in heterogeneous hyperparameter spaces~\cite{PENG2021107366, evco_a_00325}, where discrete architectural decisions interact with continuous optimization parameters (e.g., learning rates, scheduler parameters, and loss-term weights), without requiring differentiability of the outer objective. Their population-based nature also supports a practical exploration–exploitation trade-off. Rather than fully training many configurations to convergence, one can rapidly filter a large set of candidates and then concentrate computation on the most promising ones. This is especially important under fixed computational budgets. Most prior works rely on ad hoc tuning for individual problems~\citep{Buzaev2023}. Lacking systematic strategies, they may waste substantial resources on configurations that reach plateaus early.

We propose a two-phase approach that integrates global optimization methods into PINN training. First, we evaluate the quality of solutions for different hyperparameter sets over a small number of epochs and select the most promising configurations for further training in the second phase. We assessed the performance of different evolutionary algorithms for selection, such as JADE \cite{zhang2009jade}, LSHADE \cite{7969307}, Grey Wolf \cite{mirjalili2014grey}, and Whales \cite{mirjalili2016whale}. A comparison against classical approaches, including Random Search, Grid Search, the Nelder–Mead \cite{Nelder1965} method, and Bayesian optimization, on a popular and practically significant set of equations, namely Advection, Klein–Gordon and Helmholtz equations, shows that the proposed method enables systematic discovery of improved solutions.

Thus, the main contribution is as follows:
\begin{enumerate}
    \item The proposed two-stage optimization strategy enables reliable identification of high-performing PINN configurations under strict computational budgets.
    \item We show that population-based evolutionary algorithms consistently outperform classical hyperparameter tuning methods such as Bayesian optimization, random search, and grid search on benchmark PDEs.
    \item We formulate evidence-based guidelines for optimal budget distribution between exploration and exploitation phases, demonstrating that an exploration budget of approximately 10\% of standard training achieves about 40\% average improvement of the baseline-level error value under the fixed computational budget.
\end{enumerate}

Our results provide practical guidelines for robust PINN training, reducing reliance on manual tuning and enhancing applicability to complex physical systems.

\section{Related work}

Several studies have addressed the hyperparameter sensitivity of Physics-Informed Neural Networks (PINNs) through systematic or automated tuning approaches. Grid search has been applied to explore discrete hyperparameter combinations \cite{doi:10.1016/j.cma.2021.114399}, while Bayesian optimization has gained popularity for its sample efficiency in high-dimensional spaces, particularly for problems such as the Helmholtz equation \cite{arxiv:2205.06704}. However, as the number of hyperparameters increases (including architecture, activation functions, learning rates, and loss weights), these methods become computationally prohibitive due to the expensive black-box evaluations required for each configuration.

Manual tuning across a broad range of hyperparameters has also been explored \cite{doi:10.1109/ACCESS.2022.3208103}. Such hand-crafted configurations often transfer reasonably well to similar problems within the same physical domain, serving as effective starting points but lacking generalizability and requiring significant expert effort.

Neural Architecture Search (NAS) techniques have been adapted specifically to PINNs. For instance, Auto-PINN \cite{arxiv:2205.13748,Ahmad2026} employs NAS to automatically optimize network architecture and other hyperparameters, achieving notable accuracy improvements. Similarly, NAS-PINN and related frameworks automate the search for optimal PINN designs \cite{arxiv:2305.10127}. While effective, these approaches remain computationally intensive, as they require numerous full training runs to evaluate candidate architectures.

To mitigate the high cost of full-fidelity evaluations, low-fidelity approximations—such as training networks for fewer epochs—have been proposed in the broader DNN literature \cite{arxiv:1808.05377} and references therein. These surrogate-based strategies enable faster screening of configurations during hyperparameter optimization. However, their application to PINNs, particularly in combination with NAS or evolutionary methods, remains underexplored, and practical guidelines for balancing fidelity, exploration, and final accuracy are largely absent.

Recent surveys highlight the growing use of evolutionary algorithms for PINN hyperparameter and architecture optimization \cite{arxiv:2501.06572}, motivated by their robustness in noisy, non-convex, and mixed search spaces. Nevertheless, most existing works either perform full training for every candidate or lack a structured separation between rapid global exploration and high-fidelity exploitation.

To our knowledge, this is among the first works to conduct a head-to-head comparison of multiple state-of-the-art evolutionary optimizers (including JADE, LSHADE, Grey Wolf Optimizer, and Whale Optimization Algorithm) within a practical two-stage PINN tuning pipeline. Our approach differs from prior methods (e.g. \cite{Zhang2024,Baldwin,Penwarden2022,Howard2023,Zhou2025}) in that it explicitly combines population-based global search on low-fidelity (truncated-epoch) training runs for efficient exploration with full-gradient-based refinement of only the most promising candidates in the exploitation phase. This structured allocation of computational budget enables systematic discovery of superior configurations while achieving significant time savings compared to full-training baselines or NAS-only strategies.

\section{Background}
\subsection{PINN Methodology}

PINNs solve PDEs by combining the approximation capabilities of neural networks with the physical constraints defined by PDEs and their boundary conditions.
Consider a bounded domain \(\Omega \subset \mathbb{R}^d\) with boundary \(\partial\Omega\), governed by a PDE system:

\begin{equation}
\begin{aligned}
    \mathcal{D}u(\mathbf{x}) &= s(\mathbf{x}), & \mathbf{x} &\in \Omega, \\
    \mathcal{B}u(\mathbf{x}) &= g(\mathbf{x}), & \mathbf{x} &\in \partial\Omega,
\end{aligned}
\label{eq:pde_system}
\end{equation}
where \(u: \mathbb{R}^d \to \mathbb{R}^m\) is the solution field, \(\mathcal{D}\) a differential operator, \(\mathcal{B}\) is the boundary operator,  
\(s(\mathbf{x})\) and \(g(\mathbf{x})\) are known  functions. 
To approximate the true solution \(u(\mathbf{x})\), we introduce a neural network \(\hat{u}(\mathbf{x}; \psi)\), parameterized by trainable weights and biases \(\psi\). This network serves as a surrogate for \(u(\mathbf{x})\), and its parameters \(\psi\) are optimized to satisfy both the PDE and the boundary conditions.
In particular, 
the training process involves minimizing a composite loss function that quantifies how well the neural network adheres to the physical constraints. The total loss is defined as:
\begin{equation}
L(\psi) = L_{\text{PDE}}(\psi) + L_{\text{BC}}(\psi),
\end{equation}
where \(L_{\text{PDE}}(\psi)\) is the PDE residual loss, which enforces the PDE within the domain \(\Omega\):
\begin{equation}
\mathcal{L}_{\text{PDE}} = \frac{1}{N_{\Omega}} \sum_{i=1}^{N_{\Omega}} \| \mathcal{D}\hat{u}(\mathbf{x}_i) - s(\mathbf{x}_i) \|^2 
\end{equation}
and \(L_{\text{BC}}(\psi)\) is the boundary condition loss, which enforces the boundary conditions on \(\partial \Omega\):
\begin{equation}
\mathcal{L}_{\text{bc}} = \frac{1}{N_{\partial\Omega}} \sum_{j=1}^{N_{\partial\Omega}} \| \mathcal{B}\hat{u}(\mathbf{x}_j) - g(\mathbf{x}_j) \|^2
\end{equation}
In these expressions, \(N_{\Omega}\) and  \(N_{\partial \Omega}\) represent numbers of  collocation points sampled from the interior of the domain, and from the boundary, respectively. 
The optimal parameters \(\psi\) of PINN are found by minimizing the total loss during PINN training:
\begin{equation}
\psi^* = \argmin_{\psi} L(\psi).
\end{equation}
This optimization is typically carried out using gradient-based algorithms, such as Adam or L-BFGS, which use gradients of \(L(\psi)\) with respect to \(\psi\), computed via automatic differentiation. By minimizing the loss, the neural network learns to approximate the PDE solution while adhering to the boundary conditions, effectively recasting the PDE problem as an optimization task.

\subsection{Optimization with Evolutionary Algorithms}

We study several global evolutionary optimization algorithms, each based on different principles and strategies. Below is a brief description of each algorithm.

Building on the framework presented in~\cite{Buzaev2024}, this study extends the methodological toolbox to six separate evolutionary algorithms, allowing for a comprehensive evaluation of optimization performance.

The Grey Wolf Optimizer (GWO) is a nature-inspired metaheuristic that emulates the leadership hierarchy and hunting mechanism of grey wolves in nature. The algorithm categorizes wolves into four groups, representing the social hierarchy. The hunting process involves three main steps: encircling prey, hunting, and attacking prey \citep{mirjalili2014grey}.

The Whale Optimization Algorithm (WOA) is inspired by the bubble-net hunting strategy of humpback whales. It simulates the whales' behavior of creating bubble nets to encircle prey. The algorithm includes encircling prey, bubble-net attacking method (exploitation phase), and search for prey (exploration phase) \citep{mirjalili2016whale}.

JADE is an adaptive variant of the Differential Evolution (DE) algorithm that incorporates self-adaptive control parameters and an optional external archive to enhance performance. It adjusts the mutation factor and crossover rate dynamically based on historical information, improving convergence speed and solution quality \citep{zhang2009jade, Buzaev2024}.

LSHADE is an advanced version of the Success-History Adaptive Differential Evolution (SHADE) algorithm, integrating Lévy-flight-based mechanisms to enhance exploration capabilities. It employs a historical memory of successful parameter settings and adapts the population size during the optimization process. These features enable LSHADE to perform robustly on complex, high-dimensional optimization problems \citep{Tanabe2014}.

%%%%%%%%%%%%%%%%%%%%%%%%%%%%%%%%%%%%%%%%%%%%%%%%%%%%%%%%%%%%%%%%%%%%%%%%

\section{Two-stage strategy of PINN hyperparameter optimization}
The PINN approximation \(\hat{u}(\mathbf{x}; \bm{\psi}, \bm{H})\) introduces two parameter classes:

\begin{itemize}
    \item \(\bm{\psi} \in \mathbb{R}^p\): Trainable network weights/biases
    
    \item \(h \in \mathcal{H}\): Hyperparameters (learning rates, loss weights \(\lambda_{\text{pde}}, \lambda_{\text{bc}}\), network architecture)
\end{itemize}
The training process constitutes two coupled optimization tasks:
\begin{equation}
\begin{aligned}
    \text{(Inner)} & \quad \min_{\bm{\psi}} \mathcal{L}(\bm{\psi}; h) = \lambda_{\text{pde}}\mathcal{L}_{\text{pde}} + \lambda_{\text{bc}}\mathcal{L}_{\text{bc}}, \\
    \text{(Outer)} & \quad \min_{h \in \mathcal{H}} \| \hat{u}(\mathbf{x}; \bm{\psi}^*(h), h) - u_{\text{ref}}(\mathbf{x}) \|^2,
\end{aligned}
\label{eq:bilevel_opt}
\end{equation}

where \(\bm{\psi}^*(\bm{H})\) denotes weights optimized with fixed \(\bm{H}\), and $u_{\text{ref}}(\mathbf{x})$ is a reference solution used for calibrating the hyperparameters. If a reference solution is not available, the outer optimization task may be taken as follows:
\begin{equation}
\begin{aligned}
    \text{(Outer)} & \quad \min_{\bm{H}} \mathcal{L}(\bm{\psi}; \bm{H}).
\end{aligned}
\label{eq:bilevel_opt_0}
\end{equation}
The proposed two-phase optimization framework builds upon foundational advances in automated neural network design and efficient hyperparameter tuning \cite{ElskenReview}.

The computational budget is allocated across two phases. The first phase involves an outer optimization process, referred to as the exploration phase, where multiple PINNs are trained with different hyperparameter configurations to identify the most promising one. At this stage, full training of each PINN is not necessarily required; instead, it is sufficient to estimate the potential of each configuration to yield the best performance. Once the most promising configuration is identified, the exploitation phase begins, during which the selected PINN is fully trained to completion.

The search space for PINN hyperparameters encompasses both network architecture and optimization settings. 
For architecture, variations consisted of changing activation functions, the number layers and units per layer. For optimization, we explored different scheduling schemes and learning rate ranges.

Only the parameters relevant to the chosen scheduler type
% \texttt{scheduler\_type} 
are activated during the search, which reduces
the effective dimensionality of the space and speeds up the black-box optimizer.
By activating only the parameters relevant to the selected scheduler, the
optimizer explores a lower-dimensional search space, and thus converges more efficiently.

The coupled optimization problem defined in Equation~\eqref{eq:bilevel_opt} constitutes a bilevel optimization task. While classical bilevel optimization assumes convexity and smoothness for theoretical guarantees, PINN training inherently involves non-convex loss landscapes due to the neural network's nonlinear parameterization. This renders exact gradient-based hyperparameter optimization intractable, motivating our two-phase approach that reformulates the outer loop as a black-box optimization problem.

Our framework addresses this by separating exploration and exploitation:
\begin{itemize}
    \item \textbf{Exploration phase:} Global optimizers efficiently prune suboptimal regions of $\mathcal{H}$ using limited training epochs, using population-based search to avoid local minima.
    \item \textbf{Exploitation phase:} Full training of the best configuration refines the solution, capitalizing on the neural network's local convergence properties.
\end{itemize}

%%%%%%%%%%%%%%%%%%%%%%%%%%%%%%%%%%%%%%%

\section{Numerical results}
To assess the performance of our optimization-assisted PINN framework, we consider three benchmark partial differential equations: the flow mixing problem, the Klein–Gordon equation, and the Helmholtz equation. We provide additional details for these equations in Appendix~\ref{app:sec:equations}. These problems are well-established in scientific computing and are known to present significant challenges for PINN-based solvers \cite{NEURIPS2021_df438e52}. For rigorous validation, we adopt the strategy introduced in \citep{cho2023separable}, wherein an exact solution is assumed and substituted into the governing equation. The resulting residual is treated as a source term, yielding a modified problem for which the exact solution is known by construction. This approach enables direct and quantitative evaluation of the PINN’s accuracy against a ground-truth reference.
\subsection{Experimental setup}

We used for training 1000 collocation points, 100 points for boundary condition and 100 points for initial condition.
We set the following range of hyperparameters: 
\begin{itemize}
  \item \textbf{Architecture}
        \begin{itemize}
          \item number of layers: \(2\!-\!5\)
          \item units per layer: \(16\!-\!128\)
          \item activation function \(\in\)\{\texttt{tanh}, \texttt{ReLU}, \texttt{GELU}, \texttt{SiLU}\}
        \end{itemize}

  \item \textbf{Optimiser (AdamW)}
        \begin{itemize}
          \item learning rate: \(10^{-4}\!-\!10^{-2}\) (log-uniform)
          
        \end{itemize}

  \item \textbf{Learning-rate scheduler}
        \begin{itemize}
          \item \texttt{scheduler\_type} \(\in\)
                \{\texttt{reduce\_on\_plateau},
                \texttt{cosine\_annealing}, \texttt{none}\}
          \item \emph{If} \texttt{scheduler\_type = reduce\_on\_plateau}:
                \begin{itemize}
                  \item \texttt{scheduler\_factor} \(\in [0.1,\,0.5]\)
                  \item \texttt{scheduler\_patience} \(\in \{5,\,20\}\)
                \end{itemize}
        \end{itemize}
\end{itemize}
All computational experiments were performed on a single NVIDIA Tesla V100 GPU with 32 GB of memory. Each optimization iteration (10,000 training epochs) required approximately 11 minutes on average. For 
the exploration phase, we used 200 iterations per optimization run and 5 independent random seeds per equation. A population size of 30 is used across all optimizers.

\subsection{Preliminary analysis}

\begin{figure}
    \centering
    \includegraphics[width=1.0\linewidth]{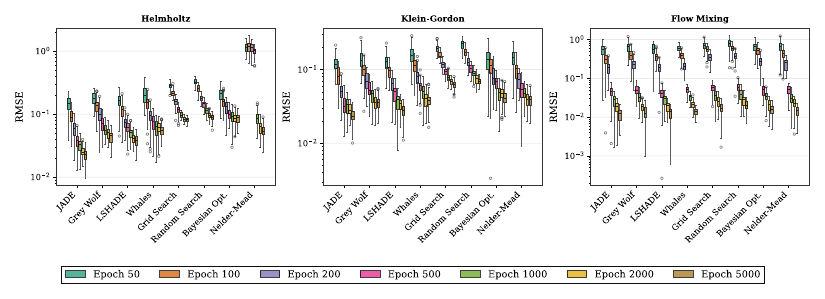}
    \caption{RMSE distribution across hyperparameter optimization methods for the Helmholtz (top), Klein-Gordon (middle), and flow mixing (bottom) equations at different epoch budgets (50--5000 epochs per trial, color-coded) with fixed total num of the iterations. Error decreases linearly (log scale) starting from 500 epochs, indicating optimal exploration budget.}
    \label{fig:fixed_iteration_boxplot}
\end{figure}

In this section, we investigate the amount of training epochs are sufficient to reliably identify the most efficient configuration at the end of training.

There is an inherent trade-off when choosing a reduced cutoff for training epochs:  
optimization algorithms must discriminate which configuration is ultimately the best,  
but waiting until full convergence is computationally expensive.  
Conversely, if training is prematurely truncated, the configuration that appears best early on may not remain the best by the end.

To study this behavior, we generated multiple configurations through random search and six global optimization algorithms.  
We analyze the resulting evaluation curves for three problems: Helmholtz3D, Klein-Gordon3D, and FlowMixing2D.

From the analysis, it is evident that to reliably identify the absolute best configuration, almost all out of 10000 epochs are typically needed (please see Appendix~\ref{app:sec:preliminary-conv} for the details).
However, requiring such a long training phase significantly increases the computational burden.  
In practice, the selection criteria may be relaxed: instead of demanding the absolute best, it is acceptable to select a configuration that finishes better than 90\% of all trials in terms of final error.

Under this relaxed criterion, it is sufficient to use approximately 200 epochs to assess and select a potentially high-performing PINN configuration.  
Early selection strategies with modest cutoff epochs (e.g., 200) can dramatically accelerate the optimization workflow without severely compromising final accuracy. In this way, the method of lower fidelity
estimates \citep{ElskenReview} and learning curve extrapolation \citep{swersky2014}
are extended to PINN.

We address the task of finding optimal architecture for the PINN model to solve a particular problem. 
\begin{figure}
    \centering
    \includegraphics[width=0.95\linewidth]{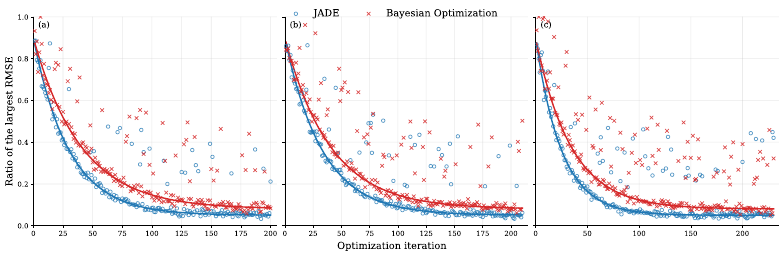}
    \caption{Saturation of the optimization process for JADE and Bayesian 
    Optimization across three benchmark PDEs: Helmholtz (top), Klein--Gordon 
    (middle), and flow mixing (bottom). The vertical axis shows the ratio 
    of the current best RMSE to the final best RMSE. Dashed vertical line 
    indicates iteration 150, where saturation is observed across all problems. }
    \label{fig:comb_error_loss}
\end{figure}

The number of trials in the exploration phase thus is driven by the nature of the original problem, and often drives the computing resources to solve the problem. 
Quality of the undertrained model for the given set of hyperparameters may be a good proxy to estimate the ultimate model quality for this set of parameters. Our hypothesis is that for PINN driven problems the rate of convergence of the model performance as a function of number of training epochs is similar for different sets of model hyperparameters. We tested this hypothesis for three PINN benchmark problems. Figure~\ref{fig:fixed_iteration_boxplot} demonstrates that indeed after a moderate number of training epochs we get a good approximation for the final model quality. 
% This is demonstrated to be valid for different optimization solvers and sets of hyperparameters. 
We found that 500 initial training epochs is a good training period to conclude, if the model with this set of hyperparameters could get to 10\% of top models.

Another important question concerns the number of iterations to perform during the exploration phase. We investigate the error reduction for varying numbers of hyperparameter optimization iterations during full retraining. As shown in Figure~\ref{fig:comb_error_loss}, after screening approximately 150 configurations additional iterations yield diminishing returns. The computational time of 32 GPU hours spent on such optimization will serve as our baseline in conditions of limited computational budget.

\subsection{Two-Stage Optimization}

\begin{figure}
    \centering
    \includegraphics[width=0.6\linewidth]{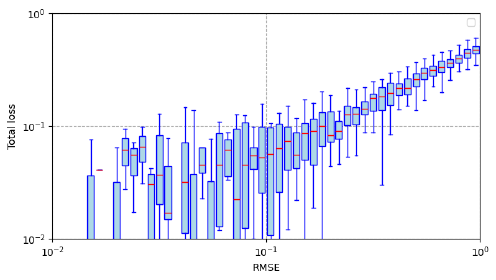}
    \caption{Relationship between the root mean square error (RMSE) and the total loss for the Helmholtz equation}
    \label{fig:error_loss}
\end{figure}

\begin{figure*}[tbh!]
    \centering
\includegraphics[width=1\linewidth]{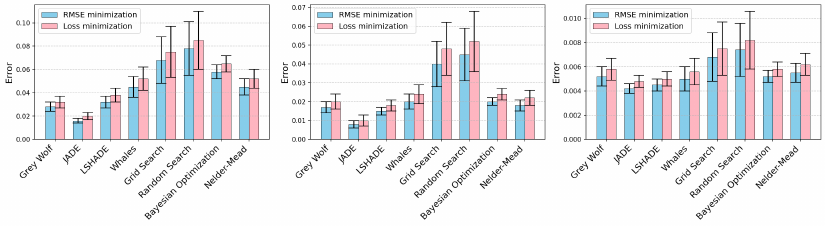}
\caption{ 
Global hyperparameter search with population-based evolutionary optimizers (Grey Wolf, JADE, LSHADE, Whales) significantly outperforms traditional baseline methods (Grid Search, Random Search, Bayesian Optimization, Nelder-Mead) across all three PDEs after 10,000 training epochs. On the Helmholtz equation (left), Klein-Gordon equation (middle), and flow mixing equation (right), evolutionary algorithms consistently achieve lower prediction errors. Results averaged over five runs; error bars show standard deviation.
\label{fig:helmholtz_comparison}
}   
\end{figure*}

We apply global optimization solvers to obtain accurate PINN solutions.
In each global optimization algorithm, we assume a fixed evaluation budget, defined as the total number of trials allowed during the exploration phase. Each
trial corresponds to training a PINN from 100 to 10000 epochs, where 10000 is the classical optimization with full learning. 
As an objective (cost) function, either the value of the total loss function or the root mean square error (RMSE) – if a reference solution is available – can be used.
Importantly, RMSE correlates with the total loss (as demonstrated in~\ref{fig:helmholtz_comparison} and Figure~\ref{fig:error_loss}),
% in Appendix~\ref{app:sec:preliminary-conv}),
which in practice allows to predict the final error based on the loss value.
Therefore, we distinguish between two optimization regimes.
When the reference solution is available, the optimization targets direct calibration of the PINN by minimizing the RMSE.
When the reference solution is not available, the optimization proceeds by minimizing the total loss function.

Figure~\ref{fig:helmholtz_comparison} shows a comparison of the efficiency of different global optimization algorithms across three problems.
Two optimization strategies are considered: (1) direct minimization of the RMSE; (2) minimization of the PINN loss function.

Although optimization based on the loss function generally leads to slightly worse results compared to RMSE minimization, it remains more practical in cases where a reference solution is not available.
For all considered equations, JADE performs best, whereas all examined evolutionary algorithms demonstrate performance that is either competitive with or superior to both Bayesian optimization and Nelder–Mead.

\begin{figure*}[tbh!]
    \centering
\includegraphics[width=1\linewidth]{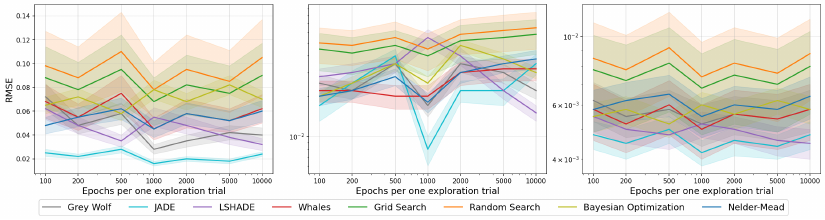}
\caption{Effect of exploration epoch budget on PINN accuracy for a fixed total hyperparameter search time across the Helmholtz equation (left), Klein-Gordon equation (middle, log scale), and flow mixing equation (right, log scale). The X-axis shows the number of epochs allocated per exploration trial (100–10,000), while the Y-axis displays the final RMSE achieved after retraining. Each curve represents one of eight global optimization algorithms. For each epoch budget, the optimizer explored the hyperparameter space using truncated training runs, then the 10 best configurations were retrained to convergence. Shaded regions indicate standard deviation across five independent runs.
\label{fig:combined_three_equations_rmse}
}   
\end{figure*}

Figure~%\ref{fig:error_per_equation} 
\ref{fig:combined_three_equations_rmse}
presents distributions of the minimal RMSE for the final trained configurations selected after conducting the exploration phase with the fixed baseline computational budget and varying number of trials and epochs (100, 200, 500, 1000, 2000, 5000, and 10000) across several optimization strategies for the Klein-Gordon equation. 
As can be seen from the Figure, for all evolutionary based optimizers, the minimum error is achieved with the number of exploratory epochs of $10^3$, which is less than full convergence. Evolutionary algorithms, in particular JADE, reach the most optimal values and outperform the Bayesian approach.

\begin{figure}[t]
    \centering
    \includegraphics[width=1.0\linewidth]{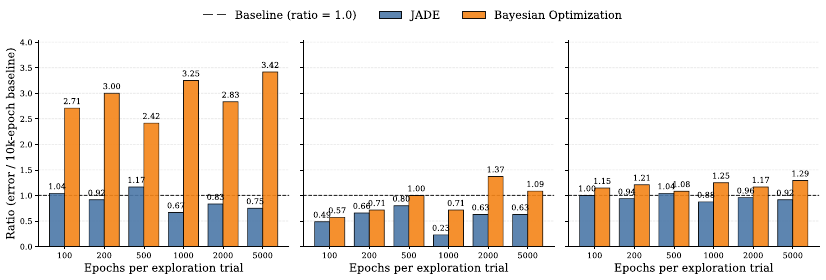}
    \caption{Computational efficiency of two-phase hyperparameter optimization for the Helmholtz equation (top), Klein-Gordon equation (middle), and flow mixing equation (bottom). The $x$-axis shows epochs per exploration trial; the $y$-axis shows the RMSE ratio to baseline (10,000 epochs, dashed line at 1.0). Values below 1.0 indicate better accuracy for the same computational budget. JADE (blue) consistently outperforms Bayesian optimization (pink), achieving 40\% average improvement at 1,000 exploration epochs, with up to 77\% error reduction for Klein-Gordon. The advantage diminishes at higher epoch budgets as both methods approach baseline performance.}
    \label{fig:jade_bayes_comparison}
\end{figure}

Figure~\ref{fig:jade_bayes_comparison} shows the computational efficiency of two-phase hyperparameter optimization for three benchmark equations. Baseline corresponds the error value of one-stage JADE optimization for full 10,000 epochs of retraining. As can be seen from the Figure, JADE optimization achieves the minimum at 1,000 epochs with error reduction of 77\% (Helmholtz), 73\% (Klein–Gordon), and 28\% (Flow Mixing) in comparison with one-stage baseline.

JADE consistently outperforms the Bayesian optimization across all tasks.
The performance gap between methods widens with increasing epochs and at 5,000 epochs JADE achieves 0.26 compared to 1.17 for Bayesian Optimization on the Helmholtz problem.

Thus, the evolutionary algorithm JADE more effectively explores the multi-modal hyperparameter landscape of PINNs due to its population-based search and adaptive adjustment of mutation parameters, whereas Bayesian Optimization becomes trapped in local optima of the surrogate model.

The best configurations and results demonstrating that the influence of individual hyperparameters is weak are shown in Appendix~\ref{app:sec:configs}.

\section{Conclusions}
In this work, we propose a two-stage framework that integrates global optimization strategies into the training of Physics-Informed Neural Networks (PINNs).
The first stage, exploration, uses evolutionary algorithms, such as JADE, LSHADE, Grey Wolf and WOA, as global optimization solvers to efficiently search for promising hyperparameter configurations with a reduced number of training epochs.
In the second stage, exploitation, the selected configuration is trained fully to convergence.

We evaluate the framework on three benchmark PDE problems, namely the Helmholtz equation, the Klein–Gordon equation, and the Flow Mixing problem.
In all cases, optimization-assisted PINNs demonstrate superior accuracy compared to baseline one-stage training strategies.
Solutions found through global optimization achieve significantly lower RMSEs and are more robust to local minima, as evidenced by improved convergence curves and more accurate reconstruction of nonmonotonic solution features. We demonstrate that RMSE correlates strongly with the total PINN loss (Figure~\ref{fig:error_loss} and \ref{fig:helmholtz_comparison}), enabling practical hyperparameter selection even when reference solutions are unavailable. The best configurations identified 
% (Table~\ref{tab:best_configurations}) show problem-specific 
architectural preferences (for instance, GELU activation for Helmholtz and flow mixing, SiLU for Klein–Gordon), though individual hyperparameter effects remain intertwined and resist simple interpretation.

Consistent with the No Free Lunch theorem \cite{Wolpert1997}, our experiments reveal no single global optimization solver that emerged as a universal solution across all considered PDE classes. However, differential evolution-based methods (with the best result for JADE) demonstrate superior stability in final solution accuracy compared to classical Bayesian approaches, random and grid search (Figure \ref{fig:helmholtz_comparison},\ref{fig:combined_three_equations_rmse} and \ref{fig:jade_bayes_comparison}). This empirical advantage likely stems from their inherent adaptation mechanisms (such as parameter self-tuning) being particularly well-suited to the structured non-convexity of PINN loss landscapes. 

Our experiments demonstrate that an exploration phase consisting of 10\% of the epochs required for full training per trial is sufficient to reliably select high-performing PINN configurations, improving model quality by 28\% to 77\% for the considered problems compared to the classical scheme with full fine-tuning per trial. Short exploration phases (about 1-2\% of the total epochs) are less reliable and require more fine-tuning, diminishing overall efficiency. Exceeding 20\% of epochs leads to diminishing returns. Thus, 10\% training epochs provide the best balance between efficiency and model quality, establishing structured hyperparameter optimization as an effective approach for resource-efficient and robust PINNs.

Future work will focus on extending the optimization framework to a broader search space that includes variations in PINN architectures and broader PDE families. 

\section*{Acknowledgments}
The work was supported by the grant for research centers in the field of AI provided by the
Ministry of Economic Development of the Russian Federation in accordance with the agreement 000000C313925P4E0002 and the agreement with HSE University 139-15-2025-009. \\
This research is supported in part through computational resources of HPC facilities at HSE University.

\bibliography{iclr2026_conference}
\bibliographystyle{iclr2026_conference}

\appendix
\section{Considered equations}
\label{app:sec:equations}

\subsection{Flow mixing problem (Advection Equation)}

We consider the two-dimensional linear advection equation:
\begin{equation}
\frac{\partial u(t,x,y)}{\partial t} + \frac{\partial u(t,x,y)}{\partial x} + \frac{\partial u(t,x,y)}{\partial y} = 0,
\end{equation}
which describes the passive transport of a scalar field in a uniform velocity field with components \( a = 1 \), \( b = 1 \) along the \( x \)- and \( y \)-axes, respectively.

An exact solution to this equation is:
\begin{equation}
u(t,x,y) = \sin\left(\pi(x - t)\right) \sin\left(\pi(y - t)\right).
\end{equation}

The computational domain is defined as \( (x,y) \in [0,1]^2 \), \( t \in [0,1] \). The initial condition is given by:
\begin{equation}
u(0,x,y) = \sin(\pi x) \sin(\pi y),
\end{equation}
which matches the exact solution at \( t = 0 \).

To ensure periodicity and compatibility with the exact solution, we impose periodic boundary conditions:
\begin{equation}
\begin{aligned}
u(t,0,y) &= u(t,1,y), \\
u(t,x,0) &= u(t,x,1),
\end{aligned}
\end{equation}
for all \( t \in [0,1] \). These conditions maintain consistency of the transported field across the boundaries throughout the simulation.

\subsection{Klein–Gordon Equation}

The Klein–Gordon equation is a nonlinear hyperbolic PDE describing relativistic scalar field dynamics. In two spatial dimensions, it takes the form:

\begin{align}
\frac{\partial^2 u(t,x,y)}{\partial t^2}
- \frac{\partial^2 u(t,x,y)}{\partial x^2}
- \frac{\partial^2 u(t,x,y)}{\partial y^2} 
&{}+ u(t,x,y)^2 \nonumber \\
&= f(t,x,y),
\end{align}
where \( (x,y) \in [-1,1]^2 \), \( t \in [0,10] \), and \( f(t,x,y) \) is a source term defined by:

\begin{equation}
f(t,x,y) = u(t,x,y)^2 - 4\,u(t,x,y).
\end{equation}

The exact solution is given by:

\begin{equation}
u_{\text{exact}}(t,x,y) = (x + y)\cos(2t) + xy\sin(2t).
\end{equation}

The initial conditions are derived from the exact solution:

\begin{equation}
u(0,x,y) = u_{\text{exact}}(0,x,y), \quad \frac{\partial u}{\partial t}(0,x,y) = \frac{\partial u_{\text{exact}}}{\partial t}(0,x,y).
\end{equation}

Similarly, Dirichlet boundary conditions are imposed using the exact solution:

\begin{equation}
u(t,\pm 1,y) = u_{\text{exact}}(t,\pm 1,y), \quad u(t,x,\pm 1) = u_{\text{exact}}(t,x,\pm 1).
\end{equation}

\subsection{Helmholtz equation}

We consider the three-dimensional Helmholtz equation:

\begin{equation}
\frac{\partial^2 u}{\partial x^2} + \frac{\partial^2 u}{\partial y^2} + \frac{\partial^2 u}{\partial z^2} + k^2 u(x,y,z) = f(x,y,z),
\end{equation}
where \( u(x,y,z) \) is the scalar field, \( k \) is the wave number, \( f(x,y,z) \) is a source term  defined as
\begin{equation}
f(x,y,z) = \left[-3\pi^2 + k^2\right] \sin(\pi x) \sin(\pi y) \sin(\pi z).
\end{equation}
Zero Dirichlet boundary conditions are imposed.
Consequently, the analytic solution is given by
\begin{equation}
u_\text{exact}(x,y,z) = \sin(\pi x) \sin(\pi y) \sin(\pi z).
\end{equation}

\section{Additional details for preliminary analysis}
\label{app:sec:preliminary-conv}

\begin{figure}[h]
    \centering
    \includegraphics[width=0.7\linewidth]{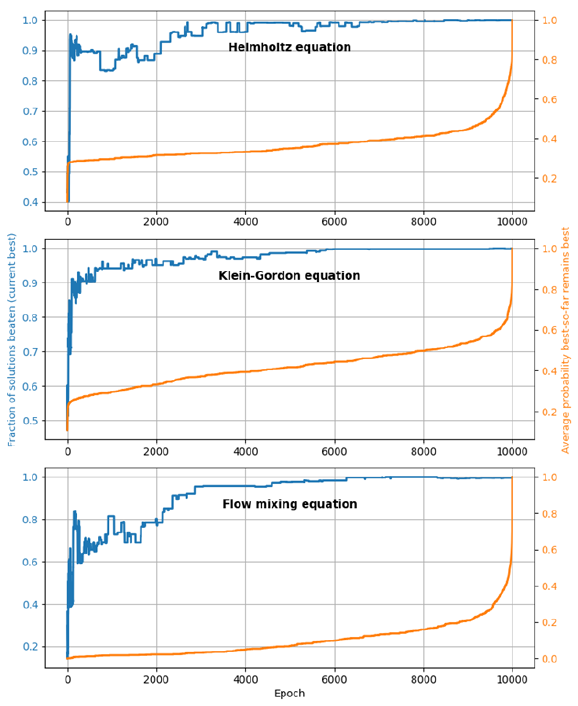}
    \caption{Prediction stability for the Helmholtz equation (top) , Klein-Gordon equation (middle) and flow mixing problem (bottom). 
        Fraction of solutions beaten by the current best configuration and average probability best-so-far remains best as function of the current epoch.  
        Results are averaged over 3000 PINN configurations.}
    \label{fig:helmholtz_quantile}
\end{figure}

The results of convergence for three equations are presented in Figures~\ref{fig:helmholtz_quantile}.
%, \ref{fig:klein_quantile}, and \ref{fig:flowmixing_quantile}.

In each figure, we plot two metrics:
\begin{itemize}
    \item The average probability best-so-far remains best (right axis, orange curve), which quantifies the likelihood that a trial currently identified as best remains the best at the final epoch.
    \item The fraction of solutions beaten (left axis, blue curve), which measures what fraction of all configurations the current best solution outperforms at the end of training.
\end{itemize}

All training processes were performed up to 10,000 epochs.  
%By examining the error curves, we observe oscillatory behavior:  
The current leading configuration may change from epoch to epoch, resulting in visible jumps in the probability curves.

\section{Configurations}
\label{app:sec:configs}

\begin{table}[ht]
\centering
\caption{Best found PINN configurations per problem for a 1000-epoch exploration phase with fixed computation time corresponding to 1279 parameter trials.}
\label{tab:best_configurations}
\begin{tabular}{l@{\hspace{1.5em}}c@{\hspace{1.5em}}c@{\hspace{1.5em}}c}
\toprule
\textbf{Parameter} & \textbf{Helmholtz} & \textbf{Klein--Gordon} & \textbf{Flow Mixing} \\
\midrule
Layers & 4 & 5 & 3 \\
Layer Size & 128 & 94 & 109 \\
Activation & GELU & SiLU & GELU \\
Learning Rate & $9.1 \times 10^{-3}$ & $4.8 \times 10^{-3}$ & $8.1 \times 10^{-3}$ \\
Scheduler & Reduce Plateau & Cosine Annealing & Reduce Plateau \\
Sch. Factor & 0.5 & 110 ($T_{\max}$) & 0.5 \\
Sch. Patience & 20 & $8.1 \times 10^{-5}$ ($\eta_{\min}$) & 16 \\
\bottomrule
\end{tabular}
\end{table}

The best configurations are presented in Table~\ref{tab:best_configurations} and demonstrates the specificity of the obtained configurations for the different equations, where the optimum is achieved with different architectures and sets of hyperparameters. 

\begin{figure}[h!]
    \centering
    \includegraphics[width=0.99\linewidth]{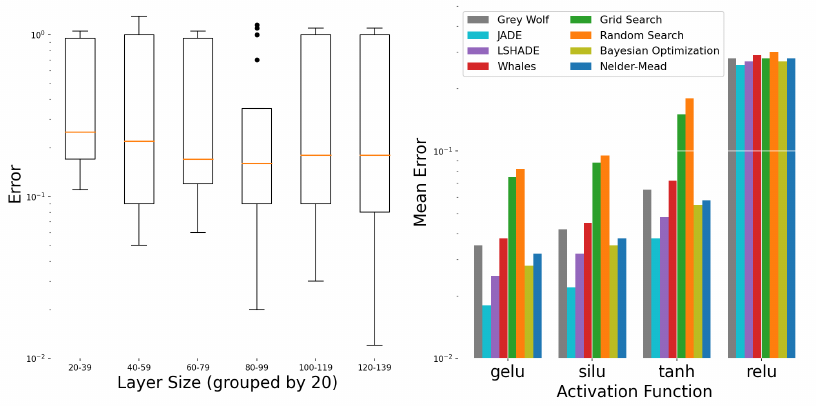}
   \caption{Mean error as a function of layer size and activation function, averaged across all considered trials and remaining hyperparameter settings for Helmholtz equation.}
    \label{fig:helmoltz_layer_act_2}
\end{figure}

The influence of individual hyperparameters is on average quite weak. The results are significantly blurred by random initialization of weights and variations in other settings. However, the activation function (see Fig.~\ref{fig:helmoltz_layer_act_2}) remains a relatively smooth and almost independent factor influencing its interpretation. At the opposite poles, the parameters of such a layer size have complex, interdependent effects that are closely intertwined with other hyperparameters. Such interrelationships complicate manual tuning and call into question the percentage of surrogate methods (e.g., Bayesian optimization), which have difficulty accurately capturing multivariate nonlinear dependencies. This, in turn, emphasizes the need for more flexible approaches.

\end{document}

%% file: math_commands.tex
\usepackage{amsmath,amsfonts,bm}

\def\eqref#1{equation~\ref{#1}}
\def\1{\bm{1}}

\DeclareMathAlphabet{\mathsfit}{\encodingdefault}{\sfdefault}{m}{sl}
\SetMathAlphabet{\mathsfit}{bold}{\encodingdefault}{\sfdefault}{bx}{n}

\DeclareMathOperator*{\argmin}{arg\,min}